**Real-Time Conflict Prediction for Large Truck Merging in Mixed Traffic at Work Zone Lane Closures**

**Abyad Enan\***
Glenn Department of Civil Engineering
Clemson University, Clemson, South Carolina, 29634
Email: aenan@clemson.edu

**Abdullah Al Mamun, Ph.D.**
Glenn Department of Civil Engineering
Clemson University, Clemson, South Carolina, 29634
Email: abdullm@clemson.edu

**Gurcan Comert, Ph.D.**
Computational Data Science and Engineering
North Carolina A&T State University, Greensboro, NC 27411
Email: gcomert@ncat.edu

**Debbie Aisiana Indah**
Department of Engineering
South Carolina State University, Orangeburg, South Carolina, 29117
Email: dindah@scsu.edu

**Judith Mwakalonge, Ph.D.**
Department of Engineering
South Carolina State University, Orangeburg, South Carolina, 29117
Email: jmwakalo@scsu.edu

**Amy W. Apon, Ph.D.**
School of Computing
Clemson University, Clemson, South Carolina, 29634
Email: aapon@clemson.edu

**Mashrur Chowdhury, Ph.D.**
Glenn Department of Civil Engineering
Clemson University, Clemson, South Carolina, 29634
Email: mac@clemson.edu

Word Count: 6,732 words + 2 tables (250 words per table) = 7,232 words

\*Corresponding Author

**ABSTRACT**
Large trucks substantially contribute to work zone-related crashes, primarily due to their large size and blind spots. When approaching a work zone, large trucks often need to merge into an adjacent lane because of lane closures caused by construction activities. This study aims to enhance the safety of large truck merging maneuvers in work zones by evaluating the risk associated with merging conflicts and establishing a decision-making strategy for merging based on this risk assessment. To predict the risk of large trucks merging into a mixed traffic stream within a work zone, a Long Short-Term Memory (LSTM) neural network is employed. For a large truck intending to merge, it is critical that the immediate downstream vehicle in the target lane maintains a minimum safe gap to facilitate a safe merging process. Once a conflict-free merging opportunity is predicted, large trucks are instructed to merge in response to the lane closure. Our LSTM-based conflict prediction method is compared against baseline approaches, which include probabilistic risk-based merging, 50th percentile gap-based merging, and 85th percentile gap-based merging strategies. The results demonstrate that our method yields a lower conflict risk, as indicated by reduced Time Exposed Time-to-Collision (TET) and Time Integrated Time-to-Collision (TIT) values relative to the baseline models. Furthermore, the findings indicate that large trucks that use our method can perform early merging while still in motion, as opposed to coming to a complete stop at the end of the current lane prior to closure, which is commonly observed with the baseline approaches.

## INTRODUCTION

The increasing demand for roadway traffic highlights the urgent need to upgrade transportation networks to meet future requirements. A critical component of these upgrades is the management of work zones, which, according to the Highway Capacity Manual, are areas of a highway where construction or repair activities affect the number of available lanes or alter traffic operations in the region (*1*). Work zones are established to support roadway construction, upgrades, or maintenance. These areas often require adjustments to normal traffic flow, such as lane closures, shifts, or reduced speed limits, creating unique challenges for both workers and drivers (*2*). The combination of constrained geometry, disrupted traffic patterns, and operational hazards can lead to crashes, injuries, and fatalities.

Large trucks, due to their size, weight, and operational characteristics, are particularly hazardous when navigating through work zones. According to the Federal Motor Carrier Safety Administration (FMCSA), large trucks refer to vehicles with a gross vehicle weight rating (GVWR) of 10,001 pounds (lbs) or more (*3*). According to the Federal Motor Carrier Safety Administration (FMCSA), large trucks were involved in 5,700 fatal crashes in 2021, marking an 18% increase compared to the previous year, alongside approximately 117,000 injury crashes and 401,000 property-damage-only crashes (*4*). Such figures highlight that large truck crashes remain a substantial concern in overall roadway safety.

Work zones, in particular, present a heightened danger for large truck operations. FMCSA reports that 6% of all fatal crashes involving large trucks occurred in work zones in 2021. Furthermore, 33% of all fatal crashes in work zones involved at least one large truck, while 15% of injury crashes in these zones also involved large trucks (*4*). According to the National Work Zone Safety Information Clearinghouse (*5*), between 2017 and 2019, over 30% of fatal work zone crashes occurred on urban interstates, while more than 55% occurred on rural interstates. At least one large truck was involved in approximately 30% of fatal work zone crashes during this period (*6*). These statistics reveal a disproportionate representation of large trucks in severe work zone incidents, underscoring the significant role they play in such crashes. Focusing on large trucks in work zone safety research is therefore crucial for multiple reasons. First, the sheer mass and momentum of large trucks increase the severity of crashes when they occur, leading to higher fatality rates and more extensive damage compared to passenger vehicles. Second, the elevated proportion of fatal work zone crashes involving large trucks suggests systemic challenges, such as limited maneuverability, reduced reaction times, and increased visibility demands, that make these vehicles more vulnerable to work zone hazards. Addressing these factors through targeted interventions, policy enhancements, and technology-assisted safety measures can help mitigate risks and reduce the alarming frequency of severe outcomes in work zone environments.

Given that roadway construction and maintenance activities are expected to rise alongside infrastructure development projects in the coming years, the implications of these statistics are clear: reducing large truck-related crashes in work zones is not only a matter of protecting truck drivers but also a critical component of enhancing overall roadway safety for all users navigating these high-risk areas (*4*).

Transportation agencies are particularly concerned about the safe navigation of large trucks in work zones, as these environments are prone to large truck-related crashes that may result in severe injuries or fatalities. Crash risk is exacerbated by the difficulty of maneuvering large vehicles through constrained, dynamically changing work zone conditions. **Figure 1** illustrates a typical work zone layout, which is divided into distinct areas, each serving a specific function in guiding and managing traffic.

Large truck lane merging typically occurs in the advance warning and transition areas, where traffic shifts and lane reductions begin. These areas are particularly challenging for large trucks due to their larger size, reduced maneuverability, and extensive blind spots. As large trucks attempt to merge within these constrained spaces, the risk of collisions with surrounding vehicles increases, especially under stop-and-go traffic conditions. In addition, drivers must remain highly alert, as changing traffic patterns and uneven pavement surfaces further complicate vehicle operation (*6*). The activity area, where construction or maintenance is actively performed, adds to the complexity as lane closures and reduced speeds further restrict traffic flow, increasing the likelihood of crashes. The termination area marks the point where traffic returns to its normal course, but the risks associated with merging persist as large trucks exit the work zone.

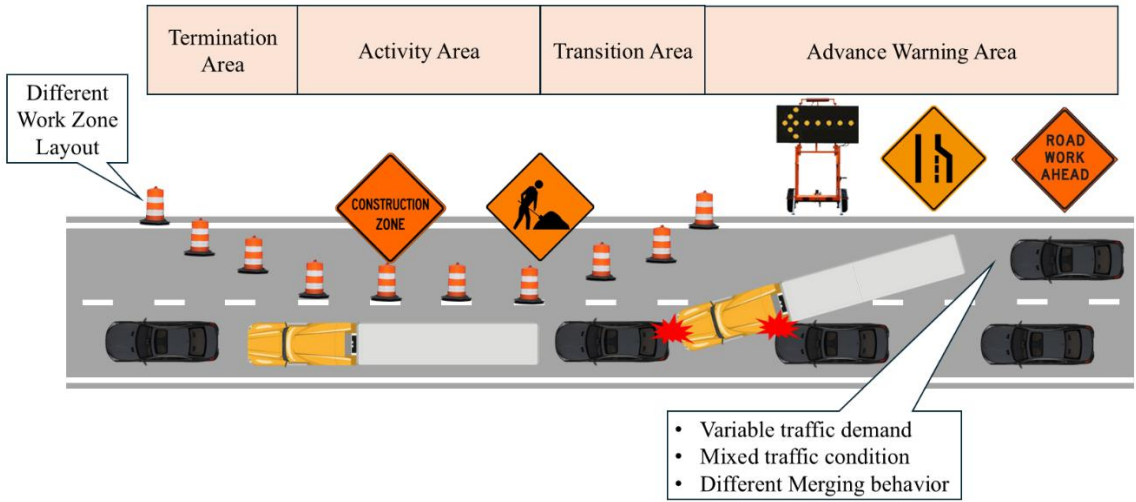

**Figure 1 Layout of a work zone showing the factors affecting large truck-related crashes**

Given these challenges, particularly during merging maneuvers in the advance warning and transition areas, the risks to both workers and other road users are substantial. Implementing a real-time conflict warning system is crucial, as it can deliver timely alerts to truck drivers, aiding in safer navigation through work zones and decreasing the chances of collisions during merging.

In this study, we develop a Long Short-Term Memory (LSTM)-based real-time conflict prediction model for large trucks, to support safe merging in the advance warning and transition areas of work zones during lane closures. The primary goal is to enhance driver awareness of surrounding traffic conditions, enabling large trucks to safely navigate through or around work zones and protect both workers and other road users.

During lane closures, vehicles are required to merge by changing lanes in the advance warning or transition areas of the work zone. In our LSTM-based conflict prediction model, the merging truck and the immediate front and rear vehicles in the target lane are considered. The model incorporates inputs such as vehicle speeds, accelerations, and relative gaps between the merging truck and adjacent vehicles. These inputs are fed into the LSTM network, which performs a binary classification to determine whether the merging event is conflicting or non-conflicting. If the model predicts a non-conflicting scenario, the truck is permitted to merge safely into the target lane and continue through the work zone. To assess the effectiveness of our model, we evaluate its performance by comparing it with baseline models in a simulated environment.

**RELATED WORKS**

Work zones are critical segments of roadways where construction, maintenance, or utility work is carried out. These zones often necessitate lane closures, which affect both traffic safety and flow efficiency. One of the primary challenges in work zones is the merging behavior of vehicles, particularly trucks, as they transition from open lanes into reduced-capacity lanes. This merging process frequently results in conflicts, especially in areas where lane closures create bottlenecks. Previous research has demonstrated that lane closures in work zones considerably reduce average roadway capacity (*7*), and increase the likelihood of merging conflicts as traffic volume remains high while available capacity diminishes. As traffic density rises, the number and severity of conflicts also tend to increase (*7*). For trucks, which are larger and slower to accelerate than passenger vehicles, the merging process is even more complex. These vehicles require larger gaps to merge safely; however, such gaps are often scarce in congested conditions. To address these challenges, various methods have been developed in prior studies to mitigate merging-related conflicts, particularly those involving heavy vehicles.

Various parametric models (*8*) have been developed to predict vehicle merging behavior, with the gap acceptance model being one of the most widely adopted approaches (*9*). In the gap acceptance model, a threshold safety gap is predetermined. Before merging, if the measured gap between the merging vehicle and both the leading and lagging vehicles exceeds this threshold safety gap, the vehicle's merging decision is triggered. However, real-world observations reveal that vehicles often change lanes even when only one of the gaps, either the lead or the lag gap, is sufficiently large, indicating that the model's assumptions do not always align with reality.

Other studies have developed logit models to predict lane-changing decisions (*8*). Weng et al. developed logistic models to determine merging decisions in work zone merging areas by considering influencing factors, such as merging vehicle and nearby vehicle types, speeds, and related gaps (*10*). Park et al. (2015) developed logistic regression models to predict discretionary lane changes in congested traffic (*10*). To increase the prediction accuracy for merging decisions, several researchers also developed non-parametric models, such as artificial neural networks and tree-based models. For example, Moridpour et al. (2012) developed a fuzzy logic lane-changing model to explain how heavy vehicles change lanes (*11*), while Meng and Weng (2012) used the classification and regression tree approach to create a non-parametric model to predict merging behavior in work zones (*12*). However, these lane-changing/merging behavior models still fail to adequately account for the time-varying effects of influencing factors. This is because real-time continuous data are needed to accurately capture driver behavior as it evolves over time.

To minimize conflict risks associated with vehicle merging, various studies have developed different strategies based on merging decisions. Marczak et al. (2013) developed a gap acceptance method for vehicle merging that uses the 50th and 85th percentile gaps as critical gap thresholds (*13*). In their method, if a merging vehicle and the nearby lead and lag vehicles in the target lane are separated by more than the critical gap distance, it is considered safe to merge. However, since these methods rely solely on the measured gap at a specific moment, they cannot be considered completely safe, as they do not account for the dynamic nature of the traffic. The accelerations and speeds of the lead or lag vehicles in the target lane are critical factors because traffic conditions can change rapidly. A gap that appears safe for merging at one moment may quickly become unsafe if the lead or lag vehicles in the target lane accelerate or decelerate. This is particularly true when merging takes more time, as even a small change in speed or distance can reduce the available space for a safe maneuver. Therefore, incorporating the speeds and accelerations of the surrounding vehicles is essential to accurately assess the safety of a merging decision over time.

Additionally, using the 85th percentile gap from naturalistic driving data can result in a large gap, and vehicles may not always find such a large gap in time to merge, resulting in late merging. In these situations, vehicles may have to come to a complete stop and wait for the expected critical gap before merging, causing operational inefficiency. To address the issue of late merging, Algomaiah et al. (2021) developed a late merging strategy for connected vehicles, where vehicles perform late merging to ensure the advance warning area of the closing lane is fully utilized under increased traffic demand conditions (*14*). However, the authors did not consider any safety metrics when evaluating the effectiveness of their method.

## CONTRIBUTIONS

In this study, we develop a real-time merging decision strategy for trucks in work zones where lane closures reduce the number of available lanes. Our method incorporates the speed and acceleration of merging trucks, as well as the kinematic information of the immediate front and rear vehicles (can be any type of vehicle) in the target lane, and the gaps between the merging truck and surrounding vehicles. These factors collectively inform the likelihood of a conflict during the merging process. By adding the kinematic data of the involved vehicles over successive time steps into a Long Short-Term Memory (LSTM) network, the model predicts whether a merging event is classified as conflicting or non-conflicting.

To evaluate the performance of our method, we conduct simulation experiments using a roadway traffic simulator under varying traffic volumes. The results are then compared with those of two baseline models. This study aims to ensure both safety and operational efficiency in real-time truck merging decisions within work zones. Our approach not only enhances safety but also facilitates early merging, which helps mitigate aggressive driving behaviors such as last-minute merges or sudden lane changes, common contributors to collisions. By addressing the limitations of previous studies, our method enables the simultaneous achievement of safety and proactive merging, thereby improving overall traffic management in work zone environments.

## LSTM-BASED CONFLICT PREDICTION METHOD

The overall truck merging process is illustrated in **Figure 2**. Initially, traffic data is collected within the work zone. When a large truck is present in a lane designated for closure due to construction, specifically within the advance warning area, its kinematic information, including speed, acceleration, and position, is recorded. Simultaneously, if there are any type of vehicles immediately in front of or behind the truck in the target (merging) lane, the kinematics of the closest front and rear vehicles are also captured. Using the time-series kinematic data, forward and rear-end conflicts are predicted through pre-trained Long Short-Term Memory (LSTM) models. The forward conflict is assessed between the merging truck and the nearest front vehicle in the target lane, while the rear-end conflict is evaluated between the truck and the closest rear vehicle in the same lane. If both predicted outcomes indicate a non-conflicting situation, the truck is permitted to safely merge into the work zone.

The LSTM-based conflict prediction model for the safe merging of trucks in work zones requires several key steps. These strategies are explained in the following subsections.

## Work Zone Network Design and Vehicle Data Collection

For a conflict prediction model focused on trucks merging in work zones, it is essential to first identify what constitutes a conflict in the naturalistic merging behavior of trucks. To achieve this, a work zone is designed using Simulation for Urban Mobility (SUMO), which is a microscopic roadway traffic simulation software. In SUMO, a network is created to replicate a real-world work zone. A two-lane, one-way roadway is reduced to a single lane due to the closure of the rightmost lane, as illustrated in **Figure 1**. The speed limit on the two-lane roadway is initially 55 miles per hour (mph) but is reduced to 50 mph within the work zone. Before the lane closure, a 3,900-feet (ft) advance warning area is marked with traffic signs indicating the lane closure and the reduced speed limit (50 mph). A smooth transition area, where the two lanes merge into a single lane (tapered), follows the advance warning area. The total length of the work zone is one mile, followed by a termination area where the work zone ends, and the roadway transitions back to two lanes with a 55-mph speed limit.

According to the Highway Capacity Manual (2010), when a two-lane roadway is reduced to a single lane in a long-term work zone, the default capacity is 1400 vehicles per hour per lane (veh/hr/ln) (*15*). In the SUMO network, vehicles are generated at this rate, with 80% of the traffic (1,120 vehicles) being passenger cars and 20% (280 vehicles) being trucks. **Table 1** summarizes the SUMO passenger car and truck dynamics of this study (*16*). The simulations are run for 10 hours, during which data is collected. During the simulation, trucks in the advance warning and transition areas are considered, particularly those about to merge. The speeds and accelerations of trucks in the rightmost lane, along with those of the vehicles immediately in front and behind in the target lane, are recorded. Additionally, the gaps between the merging trucks and the vehicles in front and behind them in the target lane are measured. Data is collected every 10 milliseconds (ms), resulting in a total of 467,003 data points.

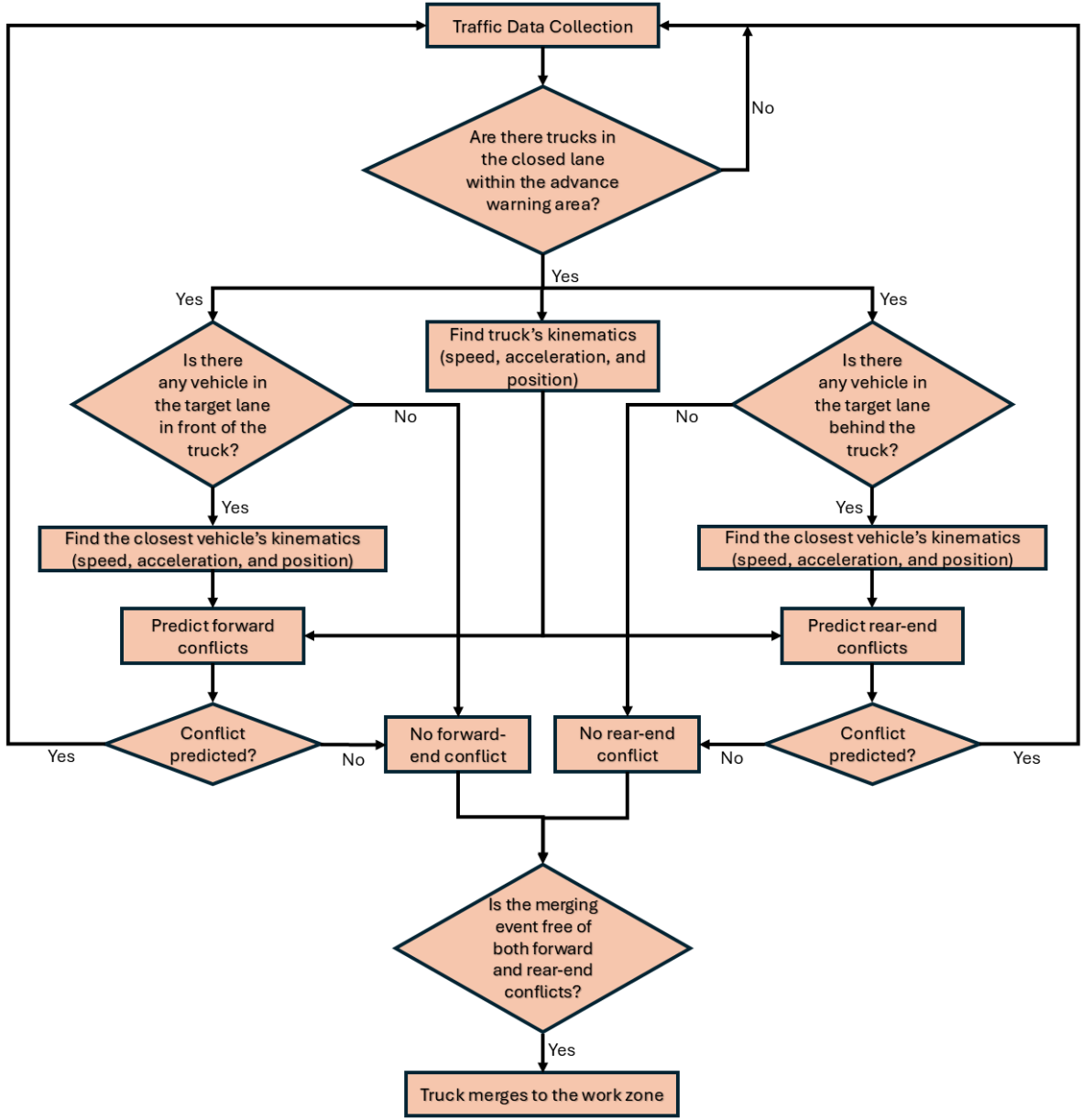

**Figure 2 LSTM-based conflict prediction-based merging strategy framework for large trucks in work zones**

**TABLE 1 Passenger Car and Truck Dynamics in SUMO (*16*) in US customary units**

| Vehicle Type | Length/width/height (ft) | Mass (lbs) | Acceleration ft/sec² | Deceleration (ft/sec²) | Maximum Speed (mph) |
|---|---|---|---|---|---|
| Passenger Car | 16.4/5.9/4.92 | 3360.9 | 8.53 | 14.76 | 124.27 |
| Large Truck | 54.13/8.37/13.12 | 28660.1 | 3.28 | 13.12 | 80.78 |

**Conflict Identification**

When a truck is about to merge, it may encounter two types of conflicts: (1) a forward conflict with the vehicle immediately in front in the target lane, and (2) a rear-end conflict with the vehicle immediately behind in the target lane. Time-to-collision (TTC) is a widely used metric for identifying conflicts (*17–19*). To identify a conflict, a threshold TTC of 2 seconds at a given time step (t) is considered, as referenced in previous studies (*17–19*). A conflict is defined using **Equation 1**:

$$conflict(t, i) = \begin{cases} 1, & 0 < TTC < 2 \ seconds \\ 0, & otherwise \end{cases} \quad (1)$$

where $i$ indicates whether it is a forward conflict (($conflict$ = 0) or a rear-end conflict (($conflict$ = 1). $t$ represents the time step at which it is determined whether a merging event is conflicting or not. TTC is calculated using **Equation 2**:

$$\frac{1}{2}\Delta a \times t^2 + \Delta V \times t - \Delta D = 0 \quad (2)$$

where, $\Delta a$, $\Delta V$, and $\Delta D$ are the relative acceleration, relative speed, and gap (i.e., the distance between the rear bumper of the front vehicle and the front bumper of the rear vehicle) between a merging truck and its immediate front or rear vehicle in the target lane.

**Data Processing**

Since trucks can encounter two types of conflicts while merging in a work zone, forward and rear-end conflicts, two separate datasets are created to train two distinct LSTM neural network models. One model is trained to predict forward conflicts, while the other focuses on rear-end conflicts. For the forward conflict prediction, the merging truck's speed and acceleration, the front vehicle's (in the target lane) speed and acceleration, and the gap between them are extracted from SUMO. As vehicle speeds and accelerations change gradually, a vehicle's current kinematics depend on its previous kinematics. To predict conflicts, the kinematics of the truck and its surrounding vehicles (trucks or passenger cars in our study) over the previous 20-time steps (2 seconds) are considered. Since the TTC is set at 2 seconds, we aim to predict conflicts based on the data from the previous 2 seconds (20-time steps). Therefore, to create a data point, the kinematic data of the truck and its immediate forward vehicle in the target lane are collected for 20 consecutive time steps from the data extracted in SUMO. Each data point is then labeled as either a conflicting or non-conflicting event using **Equations (1)** and **(2)** based on the vehicle data from the 20th time-steps. Similarly, a dataset for rear-end conflict prediction is created using the merging truck's speed and acceleration, the rear vehicle's (in the target lane) speed and acceleration, and the gap between them over 20 consecutive time steps. These data points are also labeled using **Equations (1)** and **(2)**. This process results in two time-series datasets for training forward and rear-end conflict prediction models.

**Conflict Prediction**

LSTM, a variant of Recurrent Neural Networks (RNNs), is effective for time series prediction due to its ability to capture patterns and long-range dependencies over time. Unlike traditional neural networks, an LSTM model has a unique architecture consisting of memory cells and a gating mechanism to regulate the information flow. This architecture enables the model to retrain relevant information for longer periods while discarding irrelevant details. This feature is important for analyzing time series data where future values depend on both recent and past events. LSTMs are particularly suitable for handling nonlinear relationships, which makes them robust for complex real-world problems, such as weather forecasting, stock price prediction, and anomaly detection. Furthermore, they offer more stable and accurate learning by mitigating the vanishing and exploding gradient issues that often arise in conventional RNNs during backpropagation through time (*20*). This makes LSTMs ideal for sequential data modeling, where patterns change over time and prediction accuracy is critical. In this work, we use an LSTM network for binary classification (conflicting or non-conflicting) by feeding time series data into it. The LSTM network is used to analyze time series data of 20-time steps, where each time step includes five distinctive features (truck speed and acceleration, nearby vehicle speed and acceleration, and the gap between them). Two datasets, one for forward conflicts and one for rear-end conflicts, are used to generate two pre-trained LSTM models for forward and rear-end conflict predictions.

The LSTM neural network comprises an input layer, an LSTM layer, a Dropout layer, and a Dense output layer, as shown in **Figure 3**. The input layer is structured as a three-dimensional tensor: samples, time steps, and features per time step. The sample refers to the number of data points, the time steps represent the sequential time steps (20 in this case), the LSTM model will process for each sample, and the number of features is 5 per time step. The LSTM layer is the core of the network that handles sequential data by learning long-term dependencies between the time steps. This layer consists of 50 memory cells that retain important features over long sequences. The gating mechanism, comprising input, forget, and output gates, regulates the flow of information to ensure relevant data is retained while irrelevant data is discarded. The LSTM network processes the data sequentially to encapsulate temporal dependencies between time steps. A dropout layer is added as a regularization technique to prevent overfitting during training. In this network, 20% of the neurons are randomly deactivated during training for regularization. Finally, the dense layer consists of a single neuron for binary classification, using a sigmoid activation function. The activation function maps the output to a probability where a threshold of 0.5 is applied to make a binary decision (0 means non-conflicting, and 1 means conflicting).

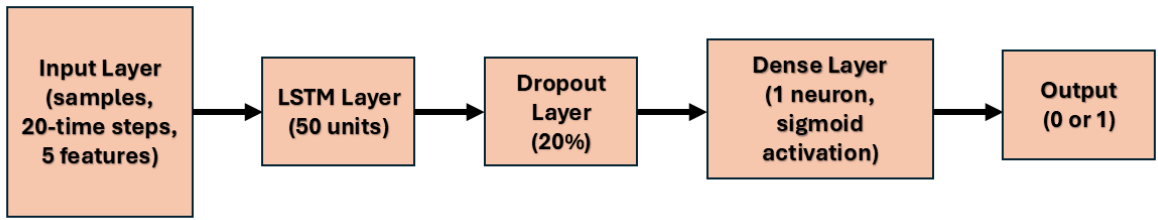

**Figure 3 LSTM network architecture**

After training this LSTM network with the two datasets, two pre-trained models are generated to predict merging conflicts (forward and rear-end). If both the forward and rear-end predictions indicate non-conflicting situations, the merging event is considered a low-risk merging, and the merging decision is triggered. If there is no vehicle ahead in the target lane, the truck only predicts rear-end conflicts. Similarly, if there is no vehicle behind in the target lane, the truck only predicts forward conflicts.

## EVALUATION STRATEGIES

Our method is evaluated by comparing its performance with baseline models. The baseline models selected for comparison are the 50th percentile gap-based merging decision method (*13*), 85th percentile gap-based merging decision method (*13*), and a probabilistic risk-based merging decision method (*21*). The evaluation strategies are explained in the following subsections.

### Experimental Setup

The work zone designed in SUMO for data collection is also used for evaluations. To assess the effectiveness of our method under varying traffic conditions, traffic volumes corresponding to 33%, 66%, and 100% of the roadway's capacity are generated to simulate different congestion scenarios. According to the Highway Capacity Manual (HCM, 2010), the default capacity for a work zone is 1,400 passenger cars per hour per lane (pc/hr/ln). For a two-lane roadway, this results in a total capacity of 2,800 pc/hr. Considering a passenger car equivalence factor of 1.5 for trucks, and a vehicle mix of 80% passenger cars and 20% trucks, the total hourly volume under full-capacity conditions comprises approximately 2,036 passenger cars and 509 trucks. For traffic volumes of 33%, 66%, and 100% capacity, the number of vehicles

generated is 840 (672 passenger cars, 168 trucks), 1,680 (1,344 passenger cars, 336 trucks), and 2,545 (2,036 passenger cars, 509 trucks) per hour, respectively.

**Features**
The two processed datasets for forward and rear-end conflicts are fitted to a Generalized Estimating Equations (GEE) model to observe the relationships between the contributing factors and their relationships with merging conflicts between the merging trucks and vehicles in the target lane for both forward and rear-end conflicts. A generalized linear model (GLM) with a potential unknown correlation between outcomes can have its parameters estimated using GEE. The GEE estimates the relationship between independent variables (truck speed, truck acceleration, vehicle speed, vehicle acceleration, measured gap) and the response variable (conflict) while considering the correlation within clusters of data. The list of variables that are used as independent variables in the GEE model to determine the relationships with forward and rear-end conflicts is provided in **Table 2**.

**TABLE 2 Estimated Contributing Factors for Forward and Rear-end Conflicts**

| Contributing Factors for Forward-end Conflict | | |
|---|---|---|
| **Variable Name** | **Coefficient** | **Standard Error** |
| Truck_speed | 0.0174 | 0.001 |
| Truck_acceleration | 0.1510 | 0.036 |
| Front_vehicle_speed | -0.0513 | 0.003 |
| Front_veh_acceleration | 0.1510 | 0.006 |
| Gap_front | -0.4209 | 0.016 |
| **Contributing Factors for Rear-end Conflict** | | |
| **Variable Name** | **Coefficient** | **Standard Error** |
| Truck_speed | -0.1690 | 0.003 |
| Truck_acceleration | -0.1048 | 0.005 |
| Rear_veh_speed | 0.1552 | 0.002 |
| Rear_veh_acceleration | 0.1520 | 0.003 |
| Gap_rear | -0.0670 | 0.002 |

**Baseline Models**

*Gap-based Models*
In the study by (*13*), the authors investigated gap-based techniques for merging decisions grounded in human driver gap acceptance principles, using two distinct critical gap values as baseline models. According to gap acceptance theory (*13*), a merging decision is made when both the forward and rear gaps, between the merging vehicle and the nearest front and rear vehicles in the target lane, exceed predefined critical thresholds. For the two gap-based baseline techniques, these critical gaps are defined as the 50th and 85th percentiles of observed gap distributions.

In our evaluation, the 50th and 85th percentile critical gaps for both the forward and rear directions are computed using data collected from the SUMO simulation. Once determined, these percentile-based thresholds are applied in the simulated work zone environment. Specifically, when a merging truck's observed forward and rear gaps are greater than or equal to the corresponding critical gap values of the selected baseline model, the truck is permitted to merge, and the event is considered a low-risk merging scenario.

*Probabilistic Risk-based Model*

In the study by (*21*), the authors developed a probabilistic risk-based merging decision method that estimates the likelihood of forward and rear-end conflicts based on gap measurements and prior event probabilities. A conflict is defined as any event where the TTC is less than or equal to 2 seconds. The method employs logistic regression to model the prior conflict probability by accounting for vehicle type in the target lane. It then estimates the probability of observing a particular gap under conflicting and non-conflicting conditions using gap distribution data. Finally, Bayes' theorem is applied to compute the posterior conflict probability given a measured gap. In our evaluation, we reimplemented this approach using SUMO-generated work zone data. Posterior probabilities for both forward and rear-end conflicts are computed at every second, and a merging decision is triggered when both posterior conflict probabilities fall below a threshold of 0.5, indicating a low-risk scenario. The details of their method can be found in (*21*). This implementation serves as another baseline to benchmark the performance of our method.

**Car-following Model**

A car-following model explains how drivers behave when navigating through traffic. The primary goal of a car-following model is to predict a vehicle's motion based on the motion of nearby vehicles. It describes how a vehicle's location, acceleration, and deceleration are affected by the movements of the surrounding vehicles. Various car-following models exist in the literature, with the Intelligent Driver Model (IDM) being one of the most popular (*22, 23*). In our SUMO simulations, the IDM car-following model is used to simulate vehicle movements. SUMO has an interactive platform called TraCI, which allows vehicle movements to be controlled (*24*). Trucks in the rightmost lane are controlled so they cannot merge into the left or target lane without receiving a merging decision or command from TraCI.

In our LSTM-based conflict prediction merging strategy, truck merging conflicts are predicted for the trucks in the rightmost lane in the advance warning area and the transition area in SUMO. Once a non-conflicting merging event is predicted, the truck receives a command to change lanes through TraCI, and the truck begins to merge into the target lane using the IDM car-following model. During and after merging, the trucks adjust their speeds, accelerations, and steering angles based on the IDM car-following model to maintain a safe distance from nearby vehicles in SUMO.

Similarly, for the gap-based baseline models, when the gaps between the trucks and the vehicles in the target lanes are greater than or equal to the threshold gaps of the respective baseline model, the merging decision is triggered by TraCI in SUMO, and the trucks merge using the IDM car-following model. Also, in the probabilistic risk-based merging decision model, if both the posterior probabilities are below the threshold (0.5), the trucks merge using the IDM car-following model.

**RESULTS AND DISCUSSIONS**

The simulation is conducted for our LSTM-based conflict prediction method for merging as well as for three baseline methods (50th percentile gap-based method, 85th percentile gap-based method, and probabilistic risk-based method) under three traffic volumes (33%, 66%, and 100% of capacity). In each case, the simulation is run for an hour, and the performance of our method is measured. For safety performance evaluation, Time Exposed Time-to-Collision (TET) and Time Integrated Time-to-Collision (TIT) are considered. For operational efficiency evaluation, early merging statistics are considered. The results are discussed in the following subsections.

**Evaluation of LSTM Model Performance for Conflict Prediction**

The LSTM network is trained using forward and rear-end conflict data from SUMO simulations to generate two pre-trained models for conflict prediction. Both processed time-series datasets are split 80%-20% for training and testing the forward and rear-end conflict prediction models. The forward conflict prediction model achieves an accuracy of 99.5%, while the rear-end conflict prediction model achieves an accuracy of 98.4%. The confusion matrices of the forward and rear-end conflict prediction models on the test data are shown in **Figure 4** below.

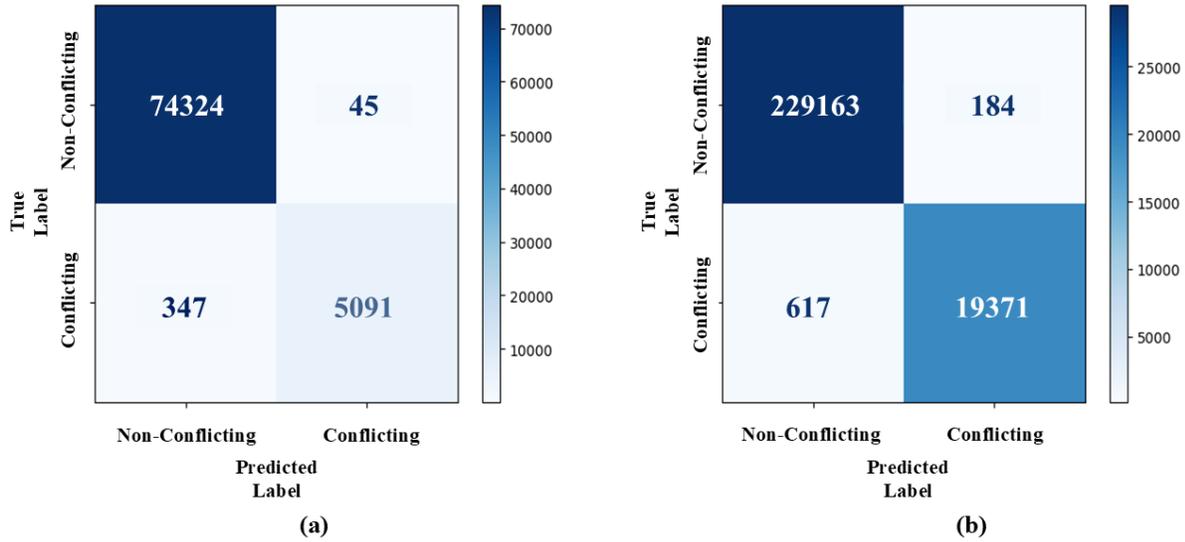

**Figure 4 Confusion matrices for LSTM-based (a) forward and (b) rear-end conflict prediction models**

## Safety Performance Evaluation

### *Time Exposed Time-to-Collision (TET)*

TET is a widely used metric to identify possible conflicts between vehicles moving on a roadway. It represents the total amount of time a vehicle spends driving in a risky or dangerous situation, where the TTC is less than or equal to 2 seconds, indicating a heightened risk of collision. A lower TET suggests that vehicles are spending less time in dangerous situations, thus reflecting a safer merging process. TET can be measured by the following equation:

$$TET(t) = \sum_{n=1}^{N} I_n \Delta t , I_n = f(x) = \begin{cases} 1, & 0 < TTC_n(t) \le 2 \\ 0, & otherwise \end{cases} \qquad (7)$$

$$TET = \sum_{t=1}^{T} TET(t) \qquad (8)$$

where, $TET(t)$ is the Time Exposed Time-to-Collision at time step $t$, n is the vehicle ID for those in the merging area, $I_n$ is a variable that equals 1 when $TTC_n$ is less than or equal to 2 seconds, and $T$ is the simulation time period.

The TET for all the merging decision-based models under different traffic volumes is summarized in **Figure 5**. The total TETs are 17.08, 48.2, 465.8, and 567.6 for the LSTM-based, probabilistic conflict prediction-based, 50th percentile-based, and 85th percentile-based merging methods, respectively, across all traffic volumes. This shows that our developed method reduces merging conflict risk by 64.56%, 96.33% and 96.99% compared to the probabilistic conflict prediction-based, 50th percentile gap-based, and 85th percentile gap-based methods, respectively. These results indicate that our method outperforms the baseline methods in terms of safety under all traffic volumes.

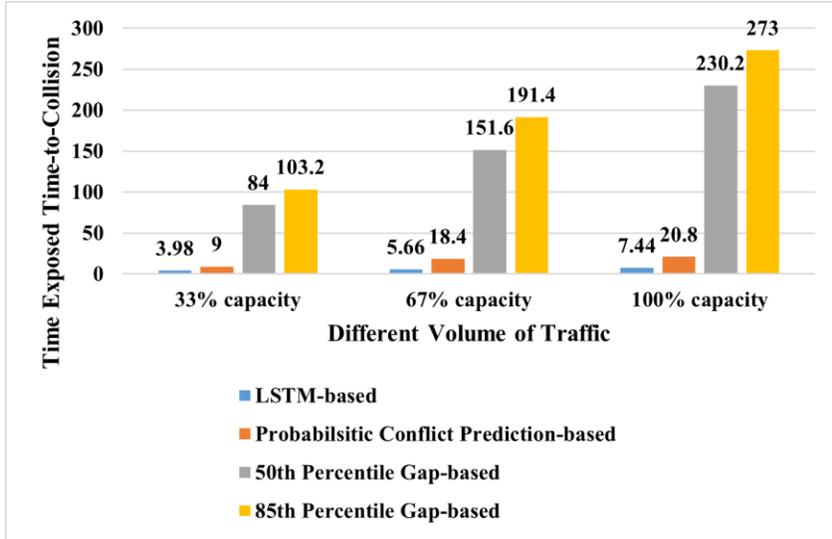

**Figure 5 Comparison of TET for different decision-based merging methods under various traffic volumes**

*Time Integrated Time-to-Collision (TIT)*

TIT measures the cumulative risk level of a vehicle by quantifying how much the TTC values fall below a safety threshold of 2 seconds. It represents the amount of time a vehicle spends in hazardous conditions during merging, where the TTC is less than 2 seconds, and how severe the conflict is based on the difference between the actual TTC and the threshold. A lower TIT value indicates that vehicles are not only spending less time in risky situations but also that the severity of the conflicts is lower. TIT can be measured using the following equations:

$$TIT(t) = \sum_{n=1}^{N}(2 - TTC_n(t)), \ 0 < TTC_n(t) \le 2 \qquad (9)$$

$$TIT = \sum_{t=1}^{T} TIT(t) \qquad (10)$$

The TIT values for all the merging decision-based models under different traffic volumes is summarized in **Figure 6**. The total TIT values are 1.87, 7.65, 55.44, and 74.23 for the LSTM-based, probabilistic conflict prediction-based, 50th percentile-based, and 85th percentile-based merging methods, respectively, across all traffic volumes. This indicates that our method reduces merging conflict risk by 75.56%, 96.63% and 97.48% compared to the probabilistic conflict prediction-based, 50th percentile-based, and 85th percentile-based merging methods, respectively.

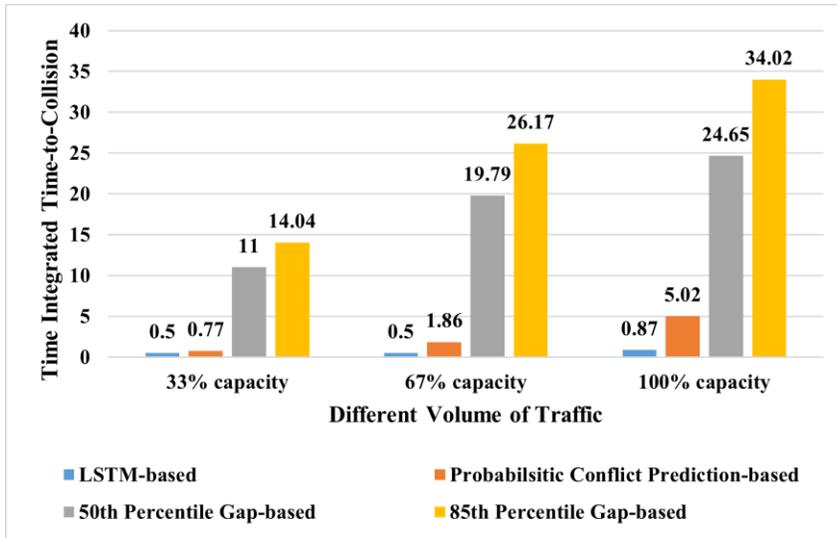

**Figure 6 Comparison of TIT for different decision-based merging methods under various traffic volumes**

**Operational Efficiency Evaluation**

To evaluate operational efficiency, we analyze the remaining distance before merging, which refers to the distance between a truck's current position and the point where the lane fully closes within the work zone, at the moment the truck begins merging. A higher average distance remains longer means the truck merges earlier in the work zone, well before reaching the lane closure point. This reduces the likelihood of trucks stopping or merging abruptly near the end of the lane, contributing to smoother traffic flow. In our evaluation, we measured the statistics of the remaining distance for our method and the baseline methods under different traffic volumes, as shown in **Figure 7**. The results show that the overall remaining distance is higher for our merging method compared to the baseline methods, suggesting that trucks in our method tend to merge earlier, improving operational efficiency.

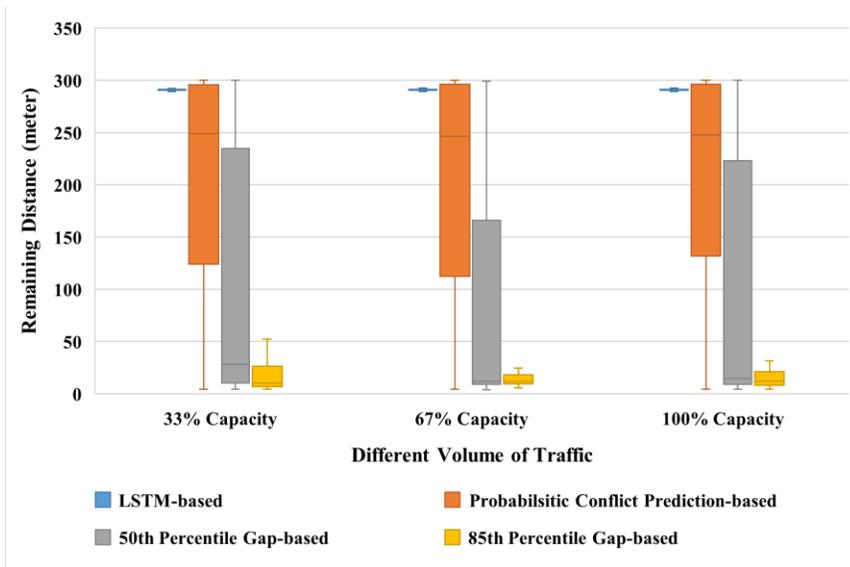

**Figure 7 Comparison of remaining merging distances for different decision-based merging methods under different traffic volumes**

## CONCLUSIONS

This study presents an LSTM-based conflict prediction strategy for real-time merging decisions of large trucks in work zones at lane closures. By incorporating large trucks and their surrounding vehicles' kinematics and relative gaps, our method predicts potential merging conflicts, which enable safer and more efficient truck merging within work zones. Through simulations under various traffic conditions in mixed traffic environments, including large trucks and passenger cars, our method demonstrated substantial improvements over baseline methods, reducing TET and TIT by up to 96.99% and 97.48%, respectively. Additionally, our method promotes earlier merging, contributing to enhanced traffic flow and reduced congestion near lane closure points.

Despite these promising results, there are some limitations to this study. The LSTM models were trained on data generated in a simulated environment due to the lack of available real-world work zone merging data for large trucks. However, our method can be adapted for real-world applications by training the LSTM models with real-world driving data. Once real-world data becomes available, the LSTM models can be fine-tuned to better reflect naturalistic driving behaviors and further improve their prediction accuracy.

Future work should prioritize the collection of real-world work zone merging data for large trucks to validate and enhance the current LSTM models. Additionally, implementing this approach in actual work zones as a pilot project could provide valuable insights into its practical utility and help refine the LSTM models further. By addressing these limitations and extending the method to real-world applications, this research could contribute significantly to improving safety and efficiency in work zone traffic management.

## ACKNOWLEDGMENTS

This work is based upon the work supported by the Federal Motor Carrier Safety Administration (FMCSA) (an agency in the United States Department of Transportation). Any opinions, findings, conclusions, and recommendations expressed in this material are those of the author(s) and do not necessarily reflect the views of FMCSA, and the U.S. Government assumes no liability for the contents or use thereof.

Note that ChatGPT was used solely to check grammar and paraphrase texts. No information, text, figure, or table has been generated, nor has any kind of analysis been conducted using any Large Language Model or Generative Artificial Intelligence.

## CONFLICT OF INTEREST

The authors declare no potential conflicts of interest with respect to the research, authorship, and/or publication of this article.

## AUTHOR CONTRIBUTIONS

The authors confirm contribution to the paper as follows: study conception and design: A. Enan, M. Chowdhury; analysis: A. Enan; interpretation of results: A. Enan, A. Mamun, M. Chowdhury, G. Comert. J. Mwakalonge, A. Apon, D. Indah; draft manuscript preparation: A. Enan, A. Mamun, M. Chowdhury, G. Comert. J. Mwakalonge, A. Apon, D. Indah. All authors reviewed the results and approved the final version of the manuscript.